\documentclass[%
 aip,
 amsmath,amssymb,
 reprint,%
]{revtex4-1}

\usepackage{graphicx}
\usepackage{dcolumn}
\usepackage{bm}

\usepackage[utf8]{inputenc}
\usepackage[T1]{fontenc}
\usepackage{mathptmx}
\usepackage{etoolbox}

\makeatletter
\def\@email#1#2{%
 \endgroup
 \patchcmd{\titleblock@produce}
  {\frontmatter@RRAPformat}
  {\frontmatter@RRAPformat{\produce@RRAP{*#1\href{mailto:#2}{#2}}}\frontmatter@RRAPformat}
  {}{}
}%
\makeatother
\begin{document}

\preprint{AIP/123-QED}

\title[Machine learning and atomic layer deposition]{Machine learning and atomic layer deposition: predicting saturation times from reactor growth profiles using artificial neural networks}
\author{Angel Yanguas-Gil}
 \email{ayg@anl.gov}
\author{Jeffrey W. Elam}%
\affiliation{ 
Applied Materials Division, Argonne National Laboratory, Lemont IL 60439 (USA)
}%

\date{\today}

\begin{abstract}
In this work we explore the application of deep neural networks to the optimization of atomic layer deposition processes based on thickness values obtained at different points of an ALD reactor. We introduce a dataset designed to train neural networks to predict saturation
times based on the dose time and thickness values measured at different points
of the reactor for a single experimental condition. We then explore different artificial neural network configurations, including depth (number of hidden layers) and size (number of neurons in each layers) to better understand the
size and complexity that neural networks should have to achieve high predictive accuracy. The results obtained show that trained neural networks can accurately
predict saturation times without requiring any prior information on the surface kinetics.
This provides a viable approach to minimize the number of experiments required
to optimize new ALD processes in a known reactor. However, the datasets and training procedure depend on the reactor geometry.

\end{abstract}

\maketitle

\section{Introduction}

Machine learning and in particular artificial neural networks have revolutionized how we 
think and work with data. While artificial neural networks are not new, and they
have long been explored as a way of modeling complex, non-linear relationships
between different variables, the development of powerful tools capable of implementing
and optimizing large networks combined with increasingly powerful computing
capabilities has lead to a veritable explosion in the range of applications and types
of architectures.

At their core, feedforward artificial neural networks are universal function approximators,
capable of modeling connections between inputs and outputs of arbitrary
complexity. This makes them
a useful tool to develop surrogate models without having to carry out complex calculations. One of the challenges, at least in the context of physical sciences,
is that they require large amounts of data for training. For instance, the MNIST dataset,
one of the most commonly used entry-level machine learning datasets, is
composed of 60,000 pictures of handwritten digits for training, plus an additional 10,000 samples for testing.\cite{Lecun_MNIST_1998}

In this work we explore the application of deep neural networks in the context of
the optimization of atomic layer deposition processes. 
In particular, we focus on a very practical question: given thickness measurements
from a set of samples distributed
inside a reactor, can we predict the dose time that would lead to saturation everywhere inside the reactor? The experimental measurement of thickness profiles using
a set of samples or witness coupons is a common approach used in research labs and in industry as part of the qualification
of a new ALD process or reactor. By training artificial neural networks to develop
a surrogate model capable of predicting saturation times from growth profiles we can
minimize the number of experiments required.

In particular, we introduce datasets designed to train neural networks to predict saturation
times based on the dose time and thickness values measured at different points
of the reactor for a single experimental condition. We then explore different artificial neural network configurations, including depth (number of hidden layers) and size (number of neurons in each layer) to better understand the
size and complexity of that neural networks should have to achieve high predictive accuracy. Finally, we evaluate the minimum number of experimental data points
required to achieve high classification accuracies. The dataset and the
code implementing the networks and the training and evaluation processes 
have been made available upon publication
of this work, and they can be found  
online at https://github.com/aldsim/saturationdataset.

\section{Methodology}

\subsection{\label{sec_dataset}Dataset}

In order to explore the application of neural networks to ALD process optimization,
we have created a simple dataset connecting growth profiles and precursor dose times
with saturation times.

As mentioned in the introduction, one of the challenges of neural networks is that
they usually require large datasets for training. This is a challenge from an experimental
standpoint. In a prior work we demonstrated that computational fluid dynamic
models provided excellent quantitative agreement with experimental growth profiles
in our own ALD reactors.\cite{YanguasGil2021}
Therefore, in this work we have used these models to
generate a large dataset that covers a representative range of experimental parameters expected in an ALD process.

For each sample in the dataset, we randomly select values for the model input parameters. These include precursor pressure, molecular mass, process temperature,
sticking probability, growth per cycle, and the corresponding dose time. We then simulate
 the reactive transport of the precursor inside the reactor both to the selected
 dose time, and we calculate the dose time required to achieve full saturation. 
 The result is a set of thickness
 values at specific locations, the dose time, and the saturation time: $(\mathbf{x}, t_d, t_\mathrm{sat})$.
 
 By repeating this process, we have constructed a series of datasets comprising 100,000
 independent samples for training, plus 10,000 independent samples for testing.
 For this work, we
 have chosen a cylindrical horizontal viscous flow reactor configuration analogous to
 the custom-built reactors in our laboratory.\cite{Elam_reactor_2002} We have created independent datasets for
 the following number of samples in the reactor (in parenthesis the separation between consecutive points): 20 (2 cm), 16 (2.5 cm), 10 (4 cm), 8 (5 cm), 5 (8 cm), and 4 (10 cm).

\subsection{Model}

We have used the datasets described in Section \ref{sec_dataset} to explore the
application of deep neural networks to learn the functional relationship:
\begin{equation}
(\mathbf{x}, t_d) \rightarrow t_\mathrm{sat}
\end{equation}
between growth profiles and dose time (the experimental observables) and the saturation dose time (our optimization target) that is central to any ALD process
optimization.

\begin{figure}
\includegraphics[width=7.5cm]{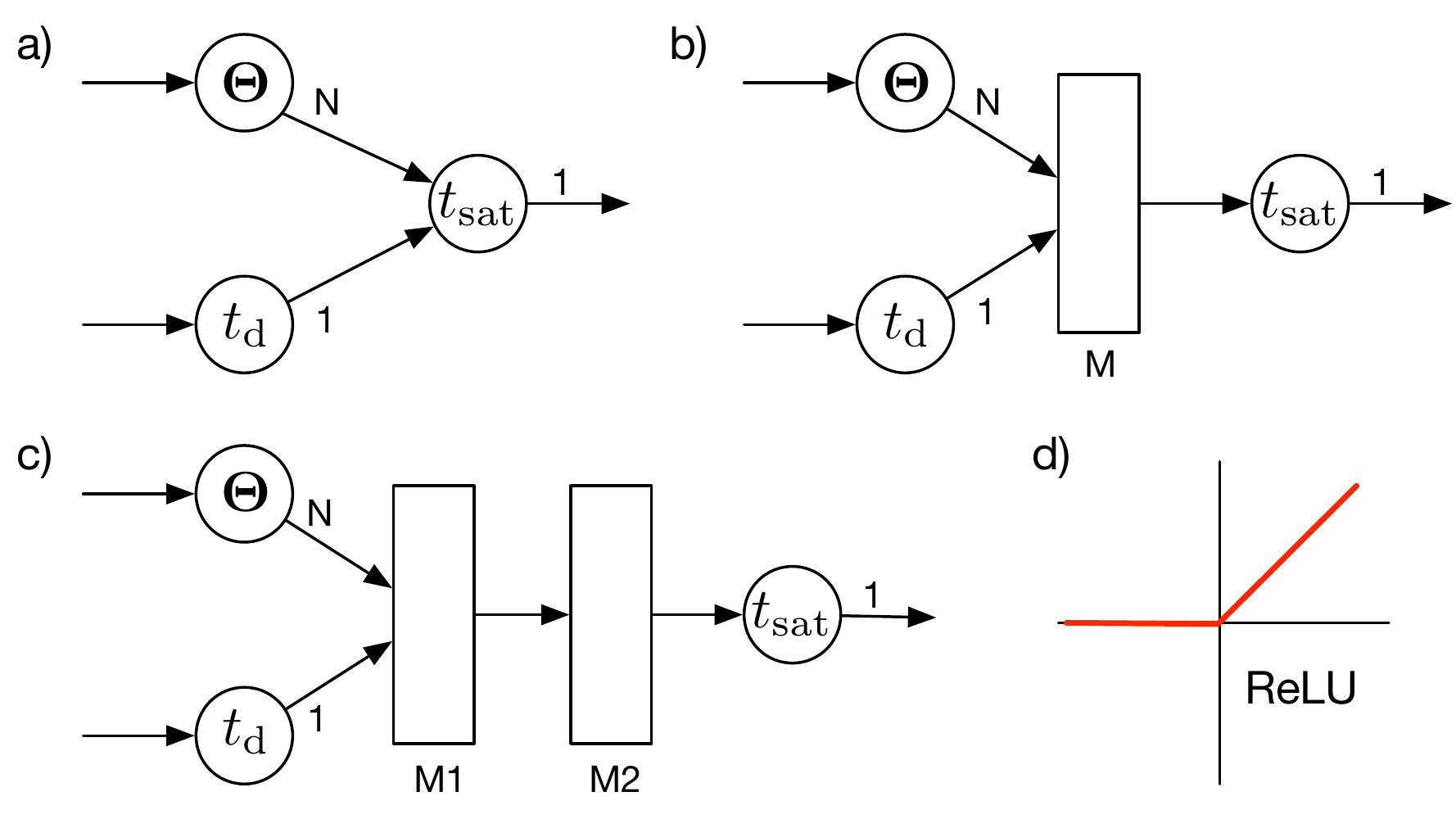}
\caption{\label{fig_scheme} a)-c) Shallow and deep neural networks considered in this work d) rectifying
linear function (ReLU) used in each of the layers.}
\end{figure}

To this end, we have explored three different networks, shown in Figure \ref{fig_scheme}:
a shallow network and two different deep 
networks with one and two hidden layers. All networks use the vector of thickness values and the dose time as input values, providing the predicted
saturation time as output. In order to achieve high accuracies over a range of times 
spanning more than two orders of magnitude, we used the logarithm of the dose and saturation times in seconds as inputs and predicted targets.

In all
cases, layers are connected
with all-to-all connections, so that the output for each layer is given by:
\begin{equation}
\mathbf{a}_i = \mathrm{ReLU}\left( \mathrm{W} \mathbf{a}_{i-1} + \mathbf{b} \right)
\end{equation}
where $\mathrm{ReLU}(\cdot)$ represents the rectified linear function [Figure \ref{fig_scheme}(d)].
One of the motivations to use all-to-all connectivities instead of convolutional layers
is that it encompasses the case where samples inside the
reactor are not equidistant from each other, but may be located downstream
or upstream to a substrate of interest. For the
deep networks, the size of each of the hidden layers, $M$ for the case of 
a single hidden layer and $M_1$ and $M_2$ for the two hidden layer
network, are additional adjustable parameters.

\subsection{\label{sec_impl} Implementation, training and testing}

We implemented the neural networks and carried out the training and
testing in Pytorch, a free open source framework
for deep learning.\cite{pytorch} Each network was trained against the training
dataset using
using stochastic gradient descent  
with a Mean Square Error (MSE) loss function and the Adam optimizer using
a learning rate of 10$^{-3}$.\cite{kingma2017adam} Each
network was trained for 100 epochs using batches of 64 samples. The
implementation and training script is provided in the Supporting Information and
can be found online at https://github.com/aldsim/saturationdataset.

The resulting networks were then tested against the testing dataset. While the
MSE was directly used as a loss function, for analysis and visualization we used
the relative difference of the predicted saturation time, defined as:
\begin{equation}
\varepsilon = \frac{t_\mathrm{pred} - t_\mathrm{sat}}{t_\mathrm{sat}}
\end{equation}
For a highly performing, unbiased network we expect this error to
have an average close to zero. The variance of $\varepsilon$, $\sigma_\varepsilon$, therefore provides a good estimator of the prediction error.

\section{Results}

\begin{figure}
\includegraphics[width=7.5cm]{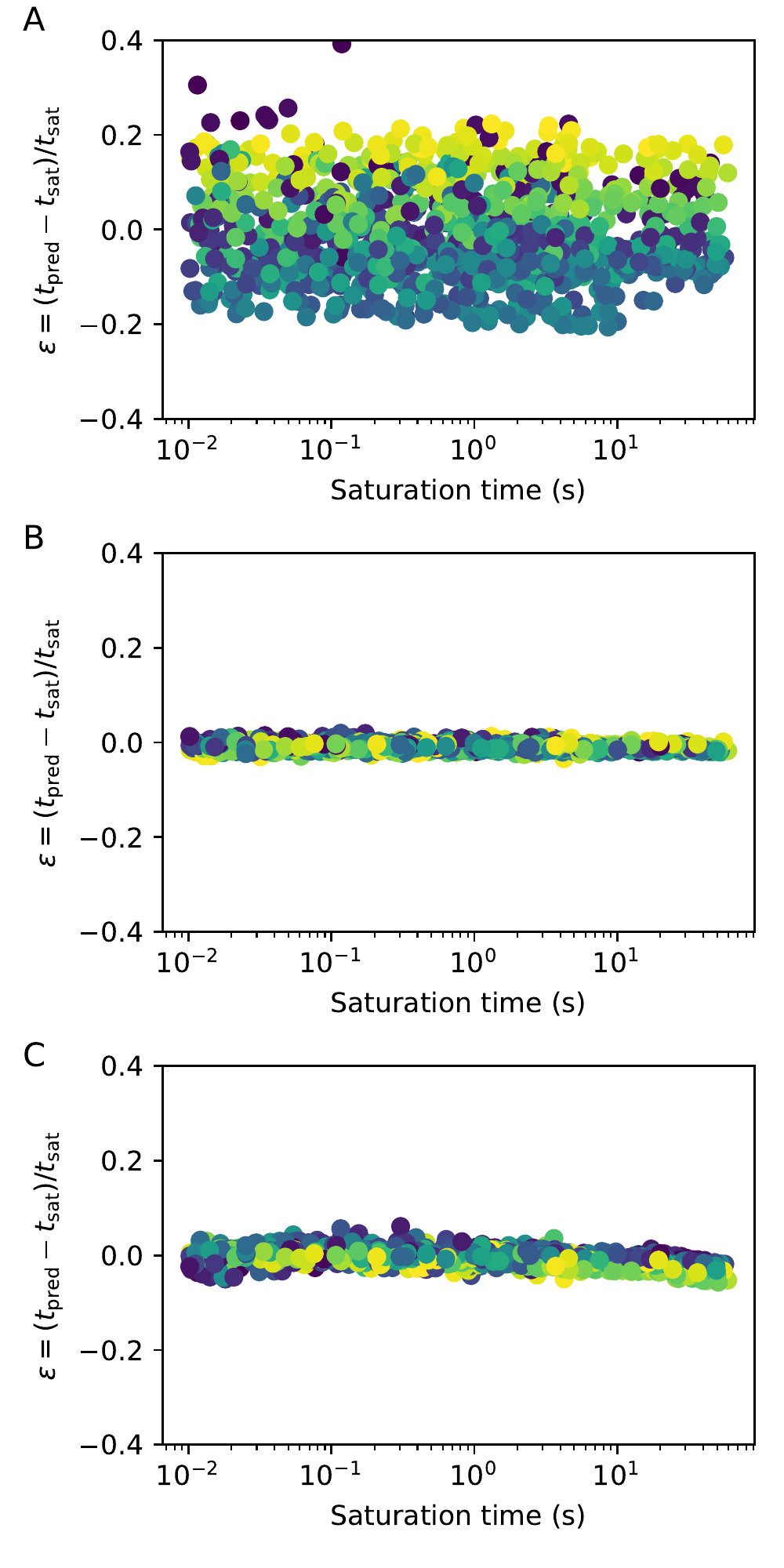}
\caption{\label{fig_error} Prediction errors in the testing dataset
for one-shot prediction of saturation
times from a single growth profile for three different neural networks:
A) Shallow network; B) 1 hidden layer ($N=30$); 
C) 2 hidden layers $N_1=30$, $N_2 = 10$. Data is shown for datasets comprising
20 independent thickness values. The data is color-coded according to how far
the dose time used for prediction is to the saturation time (lighter means shorter 
dose times)}
\end{figure}

In Figure \ref{fig_error} we show a sample of the prediction errors of three
different neural networks: a shallow network, a network with one hidden layer and a network with two hidden layers. The networks are
 trained to predict saturation times from the testing dataset
comprising 20 thickness values. Data points are colored based on how close was the dose time of the profile to the actual saturation dose, with darker points being closer
to the saturation conditions.

It is apparent that the shallow network is not capable of accurately predicting the saturation times, with the predicted times diverging $\pm20\%$  from the true
saturation value. In contrast, the deep network with one hidden layer [Fig. \ref{fig_error}(b)] shows a much smaller dispersion and an excellent agreement with the
predicted saturation times. It is interesting to note that the error seems to increase for the
deep network with two hidden layers [Fig. \ref{fig_error}(c)]. We attribute this
to an overfitting of the training dataset due to the larger number of free parameters available in the network with two hidden layers. This emphasizes the importance of having separate training and testing datasets.

As mentioned in Section \ref{sec_impl}, we can use the mean and 
standard deviation of the relative error $\varepsilon$ to quantify the network's accuracy.
In the case of the profiles shown in Figure \ref{fig_error}, the standard deviation
values are 0.10, 0.007,
and 0.015, respectively.

\subsection{Impact of network size and number of experimental points}

\begin{figure}
\includegraphics[width=7.5cm]{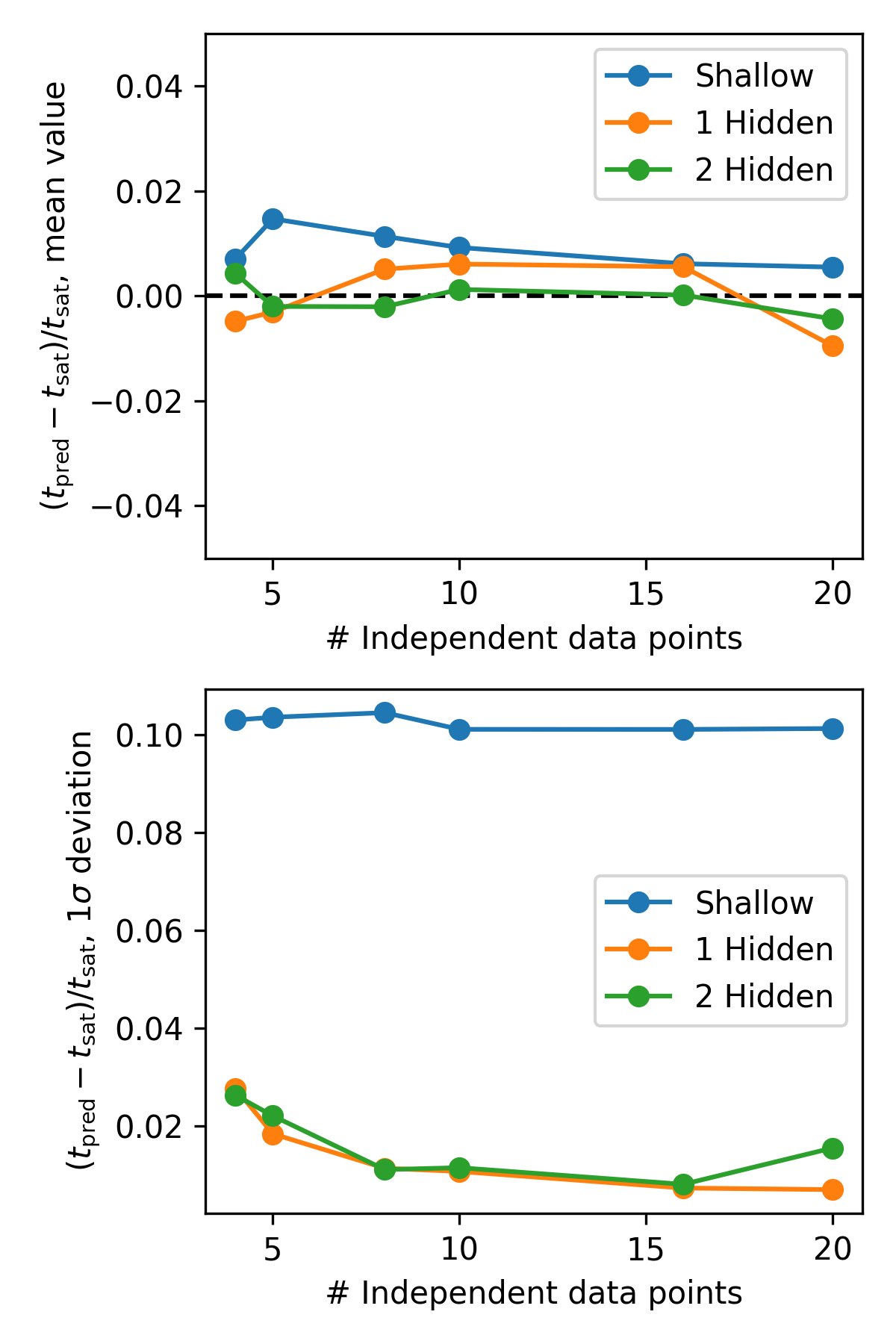}
\caption{\label{fig_npoints} Prediction accuracy as a function of the number of
independent points in the growth profile. Top: mean relative error. Bottom: standard
deviation. Results are shown for the shallow network and deep networks with 1 and 2 hidden layers.}
\end{figure}

\begin{figure}
\includegraphics[width=7.5cm]{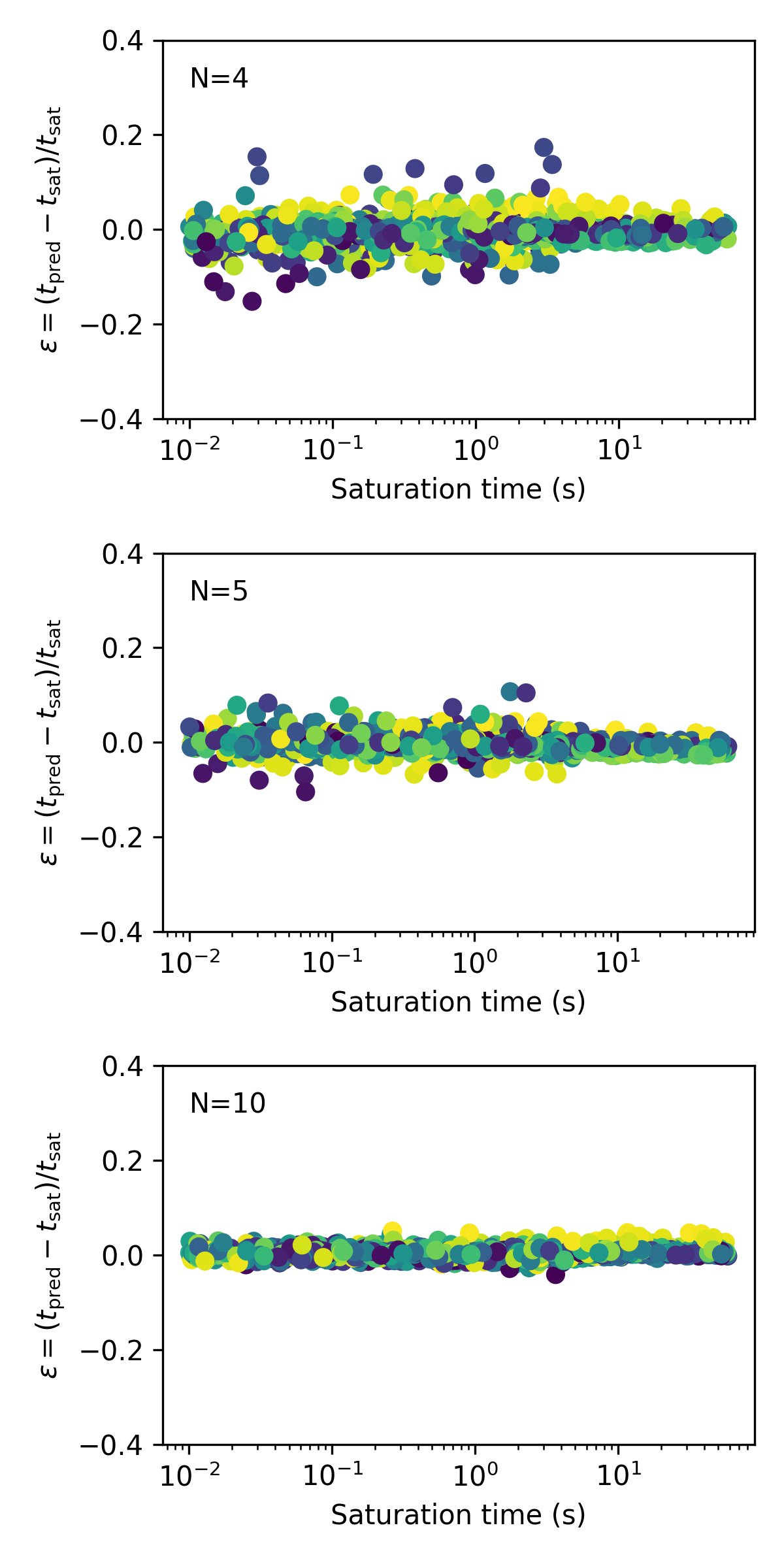}
\caption{\label{fig_points} Saturation time prediction accuracy using a 1 layer deep neural network ($M=30$) for different numbers of data points in the growth
profile: A) $N=4$, B) $N=5$ and C) $N=10$.
The data are color-coded according to how far
the dose time used for prediction is from the saturation time (lighter means shorter 
dose times)}
\end{figure}

A fundamental question when measuring growth profiles is how many experimental
points are needed to accurately capture the change in film thickness inside
a reactor. In order to understand the impact that the number of points has
on the ability to accurately predict saturation times, we trained our networks against
a collection of datasets comprising different number of homogeneously distributed points. 
In Figure \ref{fig_npoints} we show the mean error and the standard deviation $\sigma_e$ of
the predicted saturation time for the three networks shown in Figure \ref{fig_error}
when trained on different numbers of inputs. The results show that the networks with
one and two hidden layers can accurately predict the saturation time with as few as 
8 thickness values. Using fewer values still produces results that are much more accurate than those obtained using a shallow neural network, but the standard deviation in the predicted saturation times start to significantly increase. In Figure \ref{fig_points}, we show a visualization of the dispersion in the predicted
saturation values for the datasets comprising growth profiles with 4, 5, and 10 points.

\begin{figure}
\includegraphics[width=7.5cm]{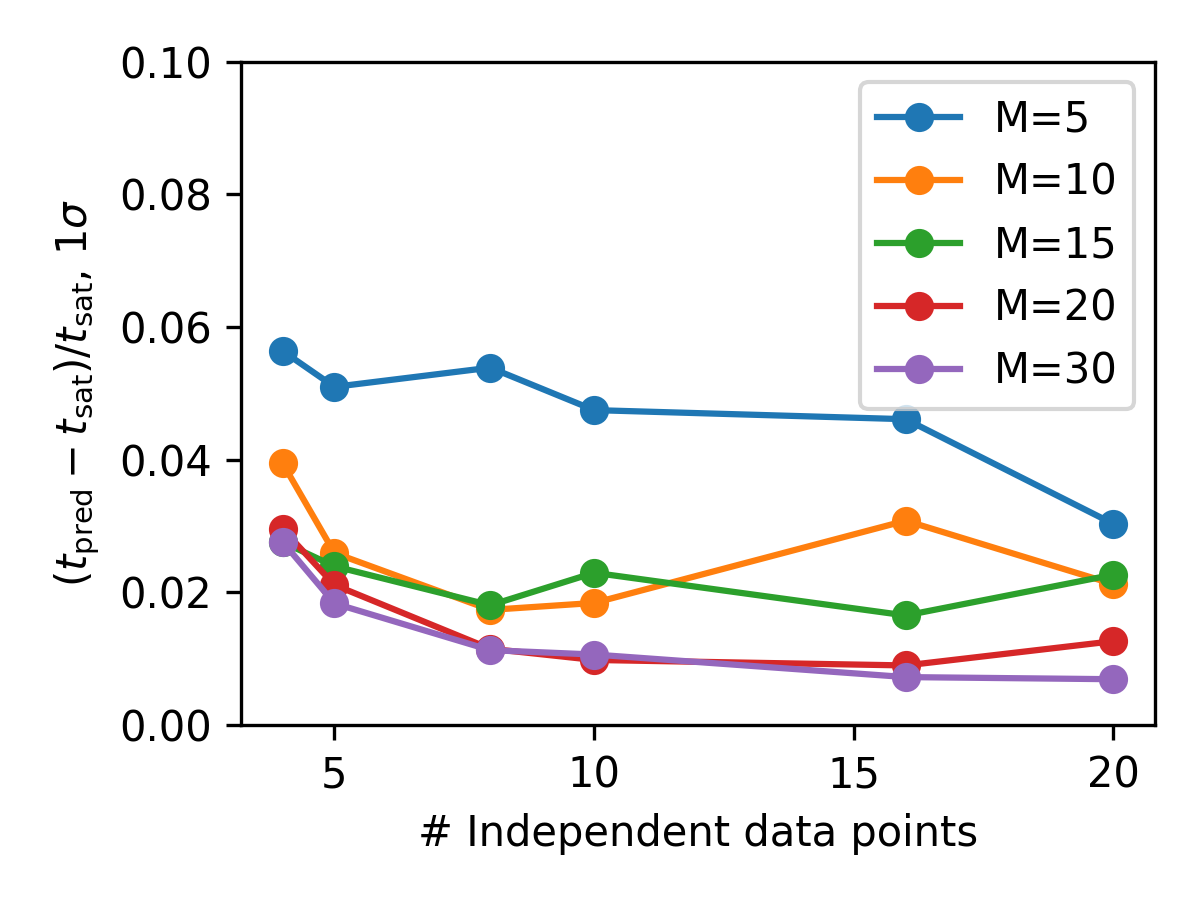}
\caption{\label{fig_size} Prediction accuracy for a network with one hidden layer as a function of the number of
independent points in the growth profile for different numbers of neurons, $M$, in the
hidden layer.}
\end{figure}

Finally, we have explored the impact of network size on its performance. In Figure \ref{fig_size}, we show the classification performance of a neural network with one hidden layer as a function of the number of independent points in the growth profile for different
hidden layer sizes. The results show how at least 20 neurons are needed in the hidden layer to maximize the network's ability to predict saturation times. This corresponds
to $21\times n + 61$ free parameters, where $n$ is the number of input data points.

\section{Conclusions}

These results demonstrate how deep neural networks can be used to predict
future behavior of reactive transport systems without any prior knowledge of the surface kineitcs, something that could help
accelerate the optimization of manufacturing processes based on thin film deposition and surface modification technologies. Compared to the networks used in traditional machine learning domains such as image classification, the number of free parameters required to achieve a good agreement is significantly smaller, 691 for the case of a
network with a single hidden layer with $N=30$ neurons.

The trained neural networks can be interpreted as surrogate models capturing the underlying physics of the reactive transport of precursors inside an ALD reactor. While the behavior far into the future of the differential equations modeling precursor transport
cannot be expressed as closed expressions, the training process is able
to capture this functional relation from the pre-existing dataset, sidestepping the
need to solve the transport models in real time.

Finally, it is important to mention that the results of the training process are reactor-specific. Consequently, specific datasets should be generated for each type of
experimental reactor. However, a single trained model can be used to predict the saturation times of arbitrary processes, including those for which no experimental information on the surface kinetics is available. Based on the results presented in this work, the accuracy seems to be limited primarily by how accurately the kinetics of a given precursor are captured in the training dataset. Therefore, departures with respect to the expected behavior could also be used to identify non-idealities in the underlying surface kinetics. 

\begin{acknowledgments}

This research is based on work supported by Laboratory Directed Research and Development (LDRD) funding from Argonne National Laboratory, provided by the Director, Office of Science, of the U.S. DOE under Contract No. DE-AC02-06CH11357.

\end{acknowledgments}

\section*{Data Availability Statement}

The data that support the findings of this study
has been made openly available in github: https://github.com/aldsim/saturationdataset.

\bibliography{aldmachine}

\begin{thebibliography}{5}%
\makeatletter
\providecommand \@ifxundefined [1]{%
 \@ifx{#1\undefined}
}%
\providecommand \@ifnum [1]{%
 \ifnum #1\expandafter \@firstoftwo
 \else \expandafter \@secondoftwo
 \fi
}%
\providecommand \@ifx [1]{%
 \ifx #1\expandafter \@firstoftwo
 \else \expandafter \@secondoftwo
 \fi
}%
\providecommand \natexlab [1]{#1}%
\providecommand \enquote  [1]{``#1''}%
\providecommand \bibnamefont  [1]{#1}%
\providecommand \bibfnamefont [1]{#1}%
\providecommand \citenamefont [1]{#1}%
\providecommand \href@noop [0]{\@secondoftwo}%
\providecommand \href [0]{\begingroup \@sanitize@url \@href}%
\providecommand \@href[1]{\@@startlink{#1}\@@href}%
\providecommand \@@href[1]{\endgroup#1\@@endlink}%
\providecommand \@sanitize@url [0]{\catcode `\\12\catcode `\$12\catcode
  `\&12\catcode `\#12\catcode `\^12\catcode `\_12\catcode `\%12\relax}%
\providecommand \@@startlink[1]{}%
\providecommand \@@endlink[0]{}%
\providecommand \url  [0]{\begingroup\@sanitize@url \@url }%
\providecommand \@url [1]{\endgroup\@href {#1}{\urlprefix }}%
\providecommand \urlprefix  [0]{URL }%
\providecommand \Eprint [0]{\href }%
\providecommand \doibase [0]{http://dx.doi.org/}%
\providecommand \selectlanguage [0]{\@gobble}%
\providecommand \bibinfo  [0]{\@secondoftwo}%
\providecommand \bibfield  [0]{\@secondoftwo}%
\providecommand \translation [1]{[#1]}%
\providecommand \BibitemOpen [0]{}%
\providecommand \bibitemStop [0]{}%
\providecommand \bibitemNoStop [0]{.\EOS\space}%
\providecommand \EOS [0]{\spacefactor3000\relax}%
\providecommand \BibitemShut  [1]{\csname bibitem#1\endcsname}%
\let\auto@bib@innerbib\@empty
\bibitem [{\citenamefont {Lecun}\ \emph {et~al.}(1998)\citenamefont {Lecun},
  \citenamefont {Bottou}, \citenamefont {Bengio},\ and\ \citenamefont
  {Haffner}}]{Lecun_MNIST_1998}%
  \BibitemOpen
  \bibfield  {author} {\bibinfo {author} {\bibfnamefont {Y.}~\bibnamefont
  {Lecun}}, \bibinfo {author} {\bibfnamefont {L.}~\bibnamefont {Bottou}},
  \bibinfo {author} {\bibfnamefont {Y.}~\bibnamefont {Bengio}}, \ and\ \bibinfo
  {author} {\bibfnamefont {P.}~\bibnamefont {Haffner}},\ }\bibfield  {title}
  {\enquote {\bibinfo {title} {Gradient-based learning applied to document
  recognition},}\ }\href {\doibase 10.1109/5.726791} {\bibfield  {journal}
  {\bibinfo  {journal} {Proceedings of the IEEE}\ }\textbf {\bibinfo {volume}
  {86}},\ \bibinfo {pages} {2278--2324} (\bibinfo {year} {1998})}\BibitemShut
  {NoStop}%
\bibitem [{\citenamefont {Yanguas-Gil}, \citenamefont {Libera},\ and\
  \citenamefont {Elam}(2021)}]{YanguasGil2021}%
  \BibitemOpen
  \bibfield  {author} {\bibinfo {author} {\bibfnamefont {A.}~\bibnamefont
  {Yanguas-Gil}}, \bibinfo {author} {\bibfnamefont {J.~A.}\ \bibnamefont
  {Libera}}, \ and\ \bibinfo {author} {\bibfnamefont {J.~W.}\ \bibnamefont
  {Elam}},\ }\bibfield  {title} {\enquote {\bibinfo {title} {Reactor scale
  simulations of ald and ale: Ideal and non-ideal self-limited processes in a
  cylindrical and a 300 mm wafer cross-flow reactor},}\ }\href {\doibase
  10.1116/6.0001212} {\bibfield  {journal} {\bibinfo  {journal} {Journal of
  Vacuum Science \& Technology A}\ }\textbf {\bibinfo {volume} {39}},\ \bibinfo
  {pages} {062404} (\bibinfo {year} {2021})},\ \Eprint
  {http://arxiv.org/abs/https://doi.org/10.1116/6.0001212}
  {https://doi.org/10.1116/6.0001212} \BibitemShut {NoStop}%
\bibitem [{\citenamefont {Elam}, \citenamefont {Groner},\ and\ \citenamefont
  {George}(2002)}]{Elam_reactor_2002}%
  \BibitemOpen
  \bibfield  {author} {\bibinfo {author} {\bibfnamefont {J.~W.}\ \bibnamefont
  {Elam}}, \bibinfo {author} {\bibfnamefont {M.~D.}\ \bibnamefont {Groner}}, \
  and\ \bibinfo {author} {\bibfnamefont {S.~M.}\ \bibnamefont {George}},\
  }\bibfield  {title} {\enquote {\bibinfo {title} {Viscous flow reactor with
  quartz crystal microbalance for thin film growth by atomic layer
  deposition},}\ }\href {\doibase 10.1063/1.1490410} {\bibfield  {journal}
  {\bibinfo  {journal} {Review of Scientific Instruments}\ }\textbf {\bibinfo
  {volume} {73}},\ \bibinfo {pages} {2981--2987} (\bibinfo {year} {2002})},\
  \Eprint {http://arxiv.org/abs/https://doi.org/10.1063/1.1490410}
  {https://doi.org/10.1063/1.1490410} \BibitemShut {NoStop}%
\bibitem [{\citenamefont {Paszke}\ \emph {et~al.}(2019)\citenamefont {Paszke},
  \citenamefont {Gross}, \citenamefont {Massa}, \citenamefont {Lerer},
  \citenamefont {Bradbury}, \citenamefont {Chanan}, \citenamefont {Killeen},
  \citenamefont {Lin}, \citenamefont {Gimelshein}, \citenamefont {Antiga},
  \citenamefont {Desmaison}, \citenamefont {Kopf}, \citenamefont {Yang},
  \citenamefont {DeVito}, \citenamefont {Raison}, \citenamefont {Tejani},
  \citenamefont {Chilamkurthy}, \citenamefont {Steiner}, \citenamefont {Fang},
  \citenamefont {Bai},\ and\ \citenamefont {Chintala}}]{pytorch}%
  \BibitemOpen
  \bibfield  {author} {\bibinfo {author} {\bibfnamefont {A.}~\bibnamefont
  {Paszke}}, \bibinfo {author} {\bibfnamefont {S.}~\bibnamefont {Gross}},
  \bibinfo {author} {\bibfnamefont {F.}~\bibnamefont {Massa}}, \bibinfo
  {author} {\bibfnamefont {A.}~\bibnamefont {Lerer}}, \bibinfo {author}
  {\bibfnamefont {J.}~\bibnamefont {Bradbury}}, \bibinfo {author}
  {\bibfnamefont {G.}~\bibnamefont {Chanan}}, \bibinfo {author} {\bibfnamefont
  {T.}~\bibnamefont {Killeen}}, \bibinfo {author} {\bibfnamefont
  {Z.}~\bibnamefont {Lin}}, \bibinfo {author} {\bibfnamefont {N.}~\bibnamefont
  {Gimelshein}}, \bibinfo {author} {\bibfnamefont {L.}~\bibnamefont {Antiga}},
  \bibinfo {author} {\bibfnamefont {A.}~\bibnamefont {Desmaison}}, \bibinfo
  {author} {\bibfnamefont {A.}~\bibnamefont {Kopf}}, \bibinfo {author}
  {\bibfnamefont {E.}~\bibnamefont {Yang}}, \bibinfo {author} {\bibfnamefont
  {Z.}~\bibnamefont {DeVito}}, \bibinfo {author} {\bibfnamefont
  {M.}~\bibnamefont {Raison}}, \bibinfo {author} {\bibfnamefont
  {A.}~\bibnamefont {Tejani}}, \bibinfo {author} {\bibfnamefont
  {S.}~\bibnamefont {Chilamkurthy}}, \bibinfo {author} {\bibfnamefont
  {B.}~\bibnamefont {Steiner}}, \bibinfo {author} {\bibfnamefont
  {L.}~\bibnamefont {Fang}}, \bibinfo {author} {\bibfnamefont {J.}~\bibnamefont
  {Bai}}, \ and\ \bibinfo {author} {\bibfnamefont {S.}~\bibnamefont
  {Chintala}},\ }\bibfield  {title} {\enquote {\bibinfo {title} {Pytorch: An
  imperative style, high-performance deep learning library},}\ }in\ \href
  {http://papers.neurips.cc/paper/9015-pytorch-an-imperative-style-high-performance-deep-learning-library.pdf}
  {\emph {\bibinfo {booktitle} {Advances in Neural Information Processing
  Systems 32}}},\ \bibinfo {editor} {edited by\ \bibinfo {editor}
  {\bibfnamefont {H.}~\bibnamefont {Wallach}}, \bibinfo {editor} {\bibfnamefont
  {H.}~\bibnamefont {Larochelle}}, \bibinfo {editor} {\bibfnamefont
  {A.}~\bibnamefont {Beygelzimer}}, \bibinfo {editor} {\bibfnamefont
  {F.}~\bibnamefont {d\textquotesingle Alch\'{e}-Buc}}, \bibinfo {editor}
  {\bibfnamefont {E.}~\bibnamefont {Fox}}, \ and\ \bibinfo {editor}
  {\bibfnamefont {R.}~\bibnamefont {Garnett}}}\ (\bibinfo  {publisher} {Curran
  Associates, Inc.},\ \bibinfo {year} {2019})\ pp.\ \bibinfo {pages}
  {8024--8035}\BibitemShut {NoStop}%
\bibitem [{\citenamefont {Kingma}\ and\ \citenamefont
  {Ba}(2017)}]{kingma2017adam}%
  \BibitemOpen
  \bibfield  {author} {\bibinfo {author} {\bibfnamefont {D.~P.}\ \bibnamefont
  {Kingma}}\ and\ \bibinfo {author} {\bibfnamefont {J.}~\bibnamefont {Ba}},\
  }\href@noop {} {\enquote {\bibinfo {title} {Adam: A method for stochastic
  optimization},}\ } (\bibinfo {year} {2017}),\ \Eprint
  {http://arxiv.org/abs/1412.6980} {arXiv:1412.6980 [cs.LG]} \BibitemShut
  {NoStop}%
\end{thebibliography}%

\end{document}